\documentclass[sigconf]{acmart}
\usepackage[ruled,vlined]{algorithm2e}
\SetKwInput{KwInput}{Input}                % Set the Input
\SetKwInput{KwOutput}{Output}              % set the Output

\usepackage{enumitem}
\usepackage{multirow} 
\newcommand{\mc}[3]{\multicolumn{#1}{#2}{#3}}
\usepackage{amsmath}
\usepackage{dsfont}
\newcommand{\eg}{{\it e.g.}}

\newcommand{\etc}{{\it etc.}}

\usepackage{placeins} % For \FloatBarrier command
\usepackage{subfigure}
\usepackage{xcolor}
\usepackage{balance}
\AtBeginDocument{%
  }

\copyrightyear{2025}
\acmYear{2025}
\setcopyright{cc}
\setcctype{by}
\acmConference[SIGIR'25]{Proceedings of the 48th International ACM SIGIR
Conference on Research and Development in Information Retrieval}{July 13--18,2025}{Padua, Italy}
\acmBooktitle{Proceedings of the 48th International ACM SIGIR Conference on
Research and Development in Information Retrieval (SIGIR'25), July 13--18,
2025, Padua, Italy}
\acmDOI{10.1145/3726302.3729958}
\acmISBN{979-8-4007-1592-1/2025/07}

\begin{document}

%%
%% The "title" command has an optional parameter,
%% allowing the author to define a "short title" to be used in page headers.
\title{Retrieval Augmented Generation for Dynamic Graph Modeling}
%\title{Dynamic Graph Modeling with Contextual and Temporal Relevance}
%%
%% The "author" command and its associated commands are used to define
%% the authors and their affiliations.
%% Of note is the shared affiliation of the first two authors, and the
%% "authornote" and "authornotemark" commands
%% used to denote shared contribution to the research.

\author{Yuxia Wu}
\affiliation{%
  \institution{Singapore Management University}
  \country{Singapore}
}
\email{yieshah2017@gmail.com}

\author{Lizi Liao}
\affiliation{%
  \institution{ Singapore Management University}
  \country{Singapore}
}
\email{lzliao@smu.edu.sg}

\author{Yuan Fang}
\authornote{Corresponding author}
\affiliation{%
  \institution{Singapore Management University}
  \country{Singapore}
}
\email{yfang@smu.edu.sg}

%%
%% By default, the full list of authors will be used in the page
%% headers. Often, this list is too long, and will overlap
%% other information printed in the page headers. This command allows
%% the author to define a more concise list
%% of authors' names for this purpose.
\renewcommand{\shortauthors}{Yuxia Wu, Lizi Liao  and Yuan Fang}

%%
%% The abstract is a short summary of the work to be presented in the
%% article.
\begin{abstract}
Modeling dynamic graphs, such as those found in social networks, recommendation systems, and e-commerce platforms, is crucial for capturing evolving relationships and delivering relevant insights over time. Traditional approaches primarily rely on graph neural networks with temporal components or sequence generation models, which often focus narrowly on the historical context of target nodes. This limitation restricts the ability to adapt to new and emerging patterns in dynamic graphs. To address this challenge, we propose a novel framework, \textbf{R}etrieval-\textbf{A}ugmented \textbf{G}eneration for \textbf{Dy}namic \textbf{G}raph modeling (\textbf{RAG4DyG}), which enhances dynamic graph predictions by incorporating contextually and temporally relevant examples from broader graph structures. Our approach includes a time- and context-aware contrastive learning module to identify high-quality demonstrations and a graph fusion strategy to effectively integrate these examples with historical contexts. The proposed framework is designed to be effective in both transductive and inductive scenarios, ensuring adaptability to previously unseen nodes and evolving graph structures. Extensive experiments across multiple real-world datasets demonstrate the effectiveness of RAG4DyG in improving predictive accuracy and adaptability for dynamic graph modeling. The code and datasets are publicly available at \url{https://github.com/YuxiaWu/RAG4DyG}.

\end{abstract}

%%
%% The code below is generated by the tool at http://dl.acm.org/ccs.cfm.
%% Please copy and paste the code instead of the example below.
%%



%%
%% Keywords. The author(s) should pick words that accurately describe
%% the work being presented. Separate the keywords with commas.
%% A "teaser" image appears between the author and affiliation
%% information and the body of the document, and typically spans the
%% page.

%\received{20 February 2007}
%\received[revised]{12 March 2009}
%\received[accepted]{5 June 2009}

%%
%% This command processes the author and affiliation and title
%% information and builds the first part of the formatted document.
\maketitle

\section{Introduction}
\begin{figure*}[ht] 
\centering
\subfigure[RAG in NLP]{
\centering
    \includegraphics[scale=0.8]{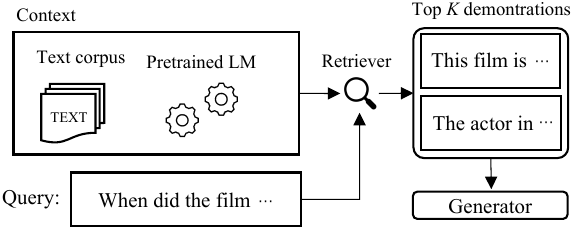}
    \label{RAG-in-NLP}
}\hspace{8mm}%
\subfigure[RAG in dynamic graphs]{
    \centering
    \includegraphics[scale=0.8]{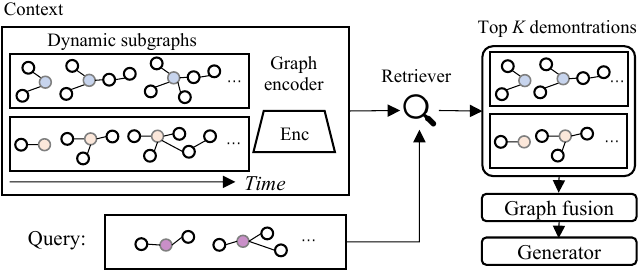}
    \label{RAG-in-Graph}
}
\vspace{-2mm}
 \caption{Illustration of RAG in NLP and dynamic graph modeling. (a) In NLP, RAG leverages pre-trained language models to encode text and retrieve semantically similar or related demonstrations, which are further concatenated to enhance the generation task. (b) Our work addresses the challenges of complex temporal and structural characteristics of dynamic graphs, incorporating RAG through time- and context-aware retrieval and graph fusion modules.}
 \vspace{-2mm}
 \label{RAG}
\end{figure*}
Dynamic graph modeling is essential for understanding and predicting evolving interactions across various applications such as social networks \cite{deng2019learning,sun2022ddgcn}, personalized recommendations\cite{song2019session, wu2022state, tang2023dynamic}, and Web-based services \cite{han2021dynamic,duan2024dga}. These applications require effective handling of the temporal and contextual dynamics inherent in evolving graph structures to provide accurate and timely insights. For example, in a recommendation system \cite{tang2023dynamic}, capturing user behavioral shifts over time can enhance the personalization of content, while in social networks \cite{sun2022ddgcn}, understanding interaction patterns can improve engagement and fraud detection mechanisms. %\lz{add some citations in this paragraph, better IR related}

Existing dynamic graph modeling methods generally fall into discrete-time and continuous-time approaches \cite{feng2024comprehensive}. Discrete-time models capture graph snapshots at specific intervals, providing a simplified representation of the graph's evolution but often failing to account for fine-grained temporal dynamics \cite{sankar2020dysat, pareja2020evolvegcn}. Continuous-time models, such as DyRep \cite{trivedi2019dyrep}, TGAT \cite{xu2020inductive}, and TREND \cite{wen2022trend}, model events as they occur, offering a more granular perspective and better capturing event-driven dynamics. These models commonly rely on graph neural networks (GNNs) integrated with temporal mechanisms such as recurrent neural networks \cite{pareja2020evolvegcn}, self-attention \cite{sankar2020dysat}, and temporal point processes \cite{wen2022trend} to update graph representations over time. Despite their strengths, GNN-based methods struggle with long-term dependencies and issues such as over-smoothing and over-squashing \cite{chen2020measuring, alon2020bottleneck}. Recently, SimpleDyG \cite{wu2024feasibility} redefined dynamic graph modeling as a generative sequence modeling task, leveraging Transformers to effectively capture long-range dependencies within temporal sequences. However, the reliance on localized historical interactions still limits the ability to generalize across different contexts and adapt to emerging patterns.

To address these limitations, the Retrieval-Augmented Generation (RAG) framework from the Natural Language Processing (NLP) domain \cite{gao2023retrieval} offers a promising approach. RAG has the potential to broaden the contextual understanding of dynamic graphs by retrieving and incorporating relevant examples from across the graph's temporal and contextual space, as illustrated in Figure \ref{RAG}. However, adopting RAG for dynamic graph modeling presents two major challenges: (1) \textit{Selecting high-quality demonstrations}, and (2) \textit{effectively integrating the retrieved demonstrations}. Identifying contextually and temporally relevant demonstrations is challenging because existing retrieval methods, such as BM25 and other matching-based schemes \cite{robertson1976relevance} primarily rely on historical interaction similarities and struggle with inductive scenarios where nodes lack historical interactions. %such as BM25  \cite{robertson1976relevance}, and text embedding-based approaches \lz{add citation?} primarily focus on semantic similarities in textual data and struggle with inductive scenarios where nodes lack historical interactions. 
Moreover, effective integration of retrieved demonstrations is another challenge, as simply concatenating them with the query sequence can lead to overly lengthy inputs and overlook underlying structural patterns.

To overcome these challenges, we propose a novel \textbf{R}etrieval-\textbf{A}ugmented \textbf{G}eneration for \textbf{Dy}namic \textbf{G}raph modeling (\textbf{RAG4DyG}) framework. It integrates a time- and context-aware contrastive learning strategy that evaluates historical interaction sequences to identify relevant examples and a graph fusion module to construct a summary graph from the retrieved samples. The contrastive learning strategy incorporates a time decay function to prioritize temporally relevant samples, while context-aware augmentation techniques such as masking and cropping enhance the model's ability to capture complex structural patterns. The graph fusion module applies a GNN-based readout mechanism to enrich the representation before feeding it into the sequence generation model. These solutions empower RAG4DyG to effectively leverage retrieved demonstrations to enhance dynamic graph modeling. Through extensive experimentation on various real-world datasets, we demonstrate that RAG4DyG outperforms state-of-the-art methods in both transductive and inductive scenarios, offering improved accuracy and adaptability in dynamic graph scenarios. In transductive settings, where test nodes have appeared during training, our model effectively leverages historical data to refine predictions. In inductive settings, involving previously unseen nodes, the retrieval mechanism enables the model to generalize by providing relevant contextual examples as guidance.

To sum up, our main contributions are as follows. (1) We propose a novel retrieval-augmented generation approach for dynamic graph modeling named RAG4DyG, which employs a retriever to broaden historical interactions with contextually and temporally relevant demonstrations. (2) We introduce a time- and context-aware contrastive learning module that incorporates temporal and structural information for demonstration retrieval and a graph fusion module to effectively integrate retrieved demonstrations. (3) We conduct extensive experiments to validate our approach, demonstrating the effectiveness of RAG4DyG across various domains.   
\iffalse
:
\begin{itemize}[topsep=0pt]
    \item We propose a novel Retrieval-Augmented Generation approach for Dynamic Graph Modeling named RAG4DyG, which employs a retriever to broaden historical interactions with contextually and temporally relevant demonstrations.
    \item We introduce a time- and context-aware contrastive learning module that incorporates temporal and structural information for demonstration retrieval and a graph fusion module for effective integration of retrieved demonstrations.
    \item We conduct extensive experiments to validate our approach, demonstrating the effectiveness of RAG4DyG across various domains.
\end{itemize}
\fi
\begin{figure*}[ht]
	\centering
	\includegraphics[scale=0.88]{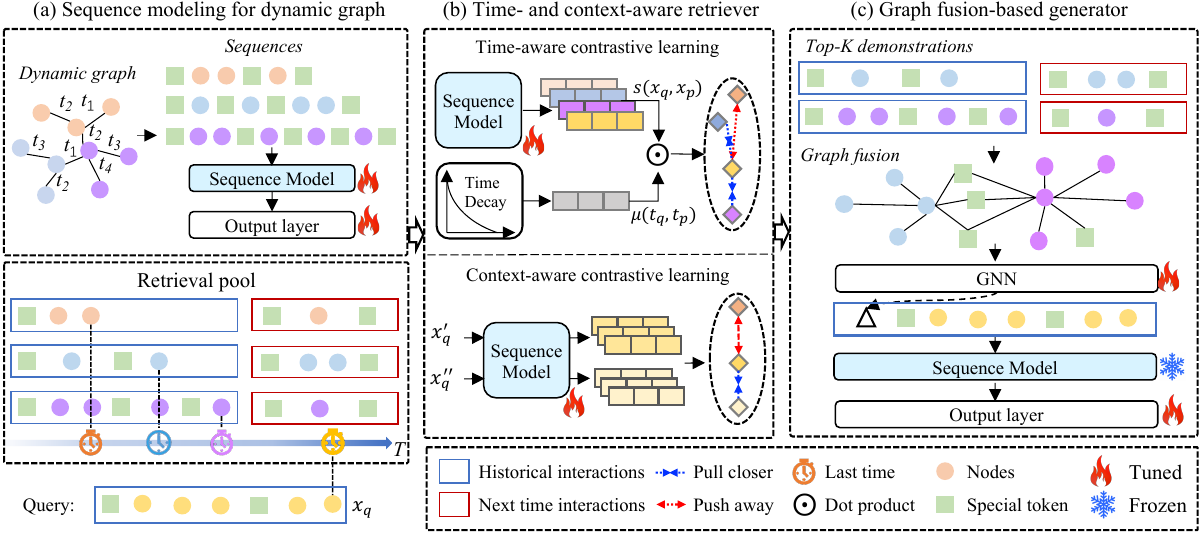}
	\caption{Overall framework of RAG4DyG. (a) Sequence modeling for dynamic graphs. (b) The retriever finds top-\textit{K} temporally and contextually relevant demonstrations. (c) Graph fusion integrates the retrieved demonstrations for the subsequent generation.}
 %(a) The dynamic graph is mapped into a series of sequences representing the interaction of ego nodes, which are processed by a sequence model. (b) A time- and context-aware retriever is designed to retrieve demonstrations for the query sample from the retrieval pool. (c) The retrieved top \textit{K} demonstrations are fused into a summary graph processed by a GNN, combined with the query sample and fed into the sequence model for link prediction.}
 
 %(a) The retrieval pool contains samples with historical and future interaction sequences, depicted in blue and red rectangles, respectively. Each interaction is associated with a specific timestamp. (b) The time-aware retriever comprises time-aware contrastive learning and self-supervised contrastive learning modules, which aim to learn better representations by pushing away dissimilar cases (red dotted lines with arrows) and pulling together similar cases (blue dotted lines with arrows). (c) The retrieval-augmented generator starts from the top \textit{K} demonstrations extracted from the retrieval pool by the retrieval model. They are composed into a summary graph followed by a GNN model, serving as enhanced context for the query example. This integrated context is then fed into a Transformer model to conduct link prediction.
 \vspace{-2mm}
\label{framework}
\end{figure*}

\section{Related Work}
\noindent\textbf{Dynamic Graph Modeling}.
Existing approaches for dynamic graphs can be categorized into discrete-time and continuous-time methods. Discrete-time methods regard a dynamic graph as a sequence of static graph snapshots captured at various time steps. Each snapshot represents the graph structure at a specific time step. These methods typically adopt GNNs to model the structural information of each snapshot, and then incorporate a sequence model \cite{pareja2020evolvegcn,sankar2020dysat} to capture the changes across snapshots. However, these approaches neglect fine-grained time information within a snapshot. %, limiting their abilities for accurate dynamic graph modeling. 
In contrast, continuous-time methods model graph evolution as a continuous process, capturing all time steps for more precise temporal modeling.
%capturing the exact time of edge formations and node interactions. 
%These methods provide more precise modeling of dynamic graphs by accounting for the continuous nature of temporal events. 
These methods often integrate GNNs with specially designed temporal modules, such as temporal random walk \cite{wang2021inductive}, temporal graph attention \cite{xu2020inductive,rossi2020temporal}, MLP-mixer \cite{cong2022we} and temporal point processes \cite{trivedi2019dyrep, ji2021dynamic, wen2022trend}. Recently, researchers have proposed a simple and effective architecture called SimpleDyG \cite{wu2024feasibility}, which reformulates dynamic graph modeling as a sequence modeling task. Specifically, it maps the dynamic graph into a series of node sequences and feeds them into a generative sequence model. Subsequently, predicting future events can be framed as a sequence generation problem. %\lz{need to further discuss the shortage of such models, and how our method differs from these} 
However, while these methods provide valuable insights, they are often limited in their ability to adapt to new and evolving patterns in dynamic graphs. %For instance, methods that integrate GNNs with temporal modules typically focus on localized temporal contexts, which may restrict their ability to capture long-term dependencies embedded in the historical data. Moreover, sequence-based models, such as SimpleDyG, rely heavily on the historical contexts of target nodes, potentially overlooking the broader contextual signals present within the dynamic graph sequences. This reliance makes it challenging for such models to adapt effectively to new patterns.

Our work distinguishes itself from prior dynamic graph learning methods through two key innovations. First, while existing approaches predominantly focus on localized temporal contexts or the historical interactions of target nodes, our proposed RAG4DyG employs retrieval-augmented generation mechanisms to retrieve and integrate broader contextual signals from the entire dynamic graph. This approach facilitates a more comprehensive understanding of dynamic interactions, uncovering complex patterns beyond the immediate historical scope of individual nodes. Second, RAG4DyG incorporates a time- and context-aware contrastive learning module for retrieving similar demonstrations, along with a graph fusion strategy to integrate them with the query sequence, enhancing adaptability to new patterns and evolving graph structures.

\vspace{+0.1cm}
\noindent\textbf{Retrieval Augmented Generation}.
Recently, the RAG paradigm has attracted increasing attention \cite{gao2023retrieval}. Specifically, RAG first leverages the retriever to search and extract relevant documents from some databases, which then serve as additional context to enhance the generation process. Related studies have demonstrated the great potential of RAG in various tasks such as language processing \cite{karpukhin2020dense, jiang2023active}, recommendation systems \cite{ye2022reflecting, dao2024broadening}, and computer vision \cite{liu2023learning, kim2024retrieval}.
In the graph modeling field, existing RAG efforts have primarily focused on static \cite{mao2023rahnet} and text-attributed graphs to enhance the generation capabilities of Large Language Models (LLMs), supporting graph-related tasks such as code summarization \cite{liuretrieval} and textual graph question answering \cite{he2024g, hu2024grag,jiang2024ragraph}. However, exploiting RAG techniques for dynamic graphs and graphs without textual information remains largely unexplored.
\section{Preliminaries}\label{sec:prelim}
In this section, we introduce the sequence modeling of dynamic graphs and the problem formulation.

\vspace{+0.1cm}\noindent\textbf{Sequence Mapping of Dynamic Graphs.}
 We denote a dynamic graph as $G=(V, E, F, \mathcal{T})$ comprising a set of nodes $V$, edges $E$, a node feature matrix $F$ if available, and a time domain $\mathcal{T}$. %Each edge is a tuple  $(v_i, v_j,t)$ representing the link between node $v_i$ and $v_j$ at time $t \in T$. 
 To map a dynamic graph into sequences, we follow SimpleDyG \cite{wu2024feasibility}. Specifically, let $D=\{(x_i, y_i)\}_{i=1}^{M}$ denote the set of training samples, where each sample is a pair $(x_i, y_i)$, representing the input and output sequences for a target node $v_i\in V$. The input $x_i$ is a chronologically ordered sequence of nodes that have historically interacted with $v_i$, while the output $y_i$ is the ground truth interactions that occur following the sequence $x_i$. In notations, we have
\begin{align}
x_i &= [\textit{hist}], v_i, [\textit{time\_1}], v_{i}^{1,1},v_{i}^{1,2}, \ldots, [\textit{time\_t}], v_{i}^{t,1}, \ldots,\nonumber\\
& \hspace{4.7mm} [\textit{time\_T}], v_{i}^{T,1}, \ldots, [\textit{eohist}], \label{eq:input}\\
y_i &= [\textit{pred}], [\textit{time\_T+1}], v_{i}^{T+1,1}, \ldots, [\textit{eopred}],\label{eq:output}
\end{align}
where $[\textit{hist}]$, $[\textit{eohist}]$, $[\textit{pred}]$, $[\textit{eopred}]$ are special tokens denoting the input and output sequence, and $[\textit{time\_1}]$, $\dots$, $[\textit{time\_T+1}]$ are special time tokens representing different time steps.

\vspace{+0.1cm}\noindent\textbf{Problem Formulation.} Dynamic graph modeling aims to learn a model that can predict the future interactions of a target node $v_i$, given its historical interactions. That is, given $x_i$ in Eq.~\eqref{eq:input}, the task is to predict $y_i$ in Eq.~\eqref{eq:output}.
In our RAG framework, we regard the 
%Compared with the existing works that heavily rely on the isolated historical context, the proposed model RAG4DyG aims to leverages the RAG technique to enhance the dynamic graph modeling. Specifically, RAG4DyG regards the 
training samples $D$ as a retrieval pool.
Given a target node $v_q \in V$, its input sequence $x_q$ is referred to as the \emph{query sequence}. We first retrieve $K$ demonstrations $R_q =  \{(x_k, y_k)\}_{k=1}^{K}$ for each query sequence $x_q$ based on their contextual and temporal relevance. Next, the retrieved demonstrations $R_q$ are used to enrich the input sequence $x_q$, which encompasses the historical interactions of the target node $v_q$. The augmented input $\{R_q, x_q\}$ is designed to enhance the predictions of future events in $y_q$.
%from $D$ to enhance the pattern understanding of the current context. Then the model integrates the valuable auxiliary information from demonstrations with the query to better learn the dynamic change of each node in the graph.

%$x_i' = \langle \text{|\textit{hist}|}\rangle, v_i, \langle \text{|\textit{time1}|}\rangle,  S_i^1, \dots \langle \text{|\textit{timeT-1}|}\rangle, S_i^{T-1}, \langle \text{|\textit{endofhist}|}\rangle,$
%Followed by the setting of SimpleDyG \cite{wu2024feasibility}, the training examples is defined as $D=\{(x_i, y_i)\}_{i=1}^{|V|}$ consists of a set of tuples $(x_i, y_i)$ denoting the historical and the corresponding future linked edges.

\section{Proposed Model: RAG4DyG}

%In this section, we first introduce an overview of the proposed RAG4DyG in Fig.~\ref{framework}. Then, we introduce the retriever and fusion modules in the following parts.

%\subsection{Overall Framework}
%Our framework 
%It first trains a generative sequence model for dynamic graphs, as shown in  Fig.~\ref{framework}(a). 
%In this work, we adopt SimpleDyG \cite{wu2024feasibility} as our backbone for sequence modeling.
%As explained in Sect.~\ref{sec:prelim}, a dynamic graph $G$ is mapped into a series of sequences representing the interaction records of each node. A Transformer-based sequence model with an output layer is then trained on the mapped sequences. 

%Our proposed RAG4DyG framework, depicted in Fig.~\ref{framework}, first trains a generative sequence model for dynamic graphs in Fig.~\ref{framework}(a). We use SimpleDyG \cite{wu2024feasibility} as the backbone for this task. As described in Sec.~\ref{sec:prelim}, a dynamic graph $G$ is represented as sequences of node interaction records, which are then used to train a Transformer-based sequence model.
The RAG4DyG framework enhances dynamic graph modeling by incorporating retrieval-augmented generation techniques to improve predictive accuracy and adaptability. As illustrated in Fig.\ref{framework}, it first adopts SimpleDyG \cite{wu2024feasibility} to model dynamic graphs as sequences of node interactions, leveraging a Transformer-based model to capture temporal dependencies and predict future interactions (Fig.\ref{framework}(a)). To enrich the modeling process, a time- and context-aware retriever retrieves relevant demonstrations from a retrieval pool $D$ for a given query sequence $x_q$. This retriever optimizes two contrastive objectives: a time-aware loss, which employs a temporal decay function $\mu(t_q, t_p)$ to prioritize temporally relevant samples, and a context-aware loss, which utilizes sequence augmentation techniques such as masking and cropping to capture structural patterns (Fig.\ref{framework}(b), Sec.\ref{sec:retriever}). Once the top-$K$ demonstrations are retrieved, they are fused into a summary graph $G_{fus}$, which captures the underlying structural relationships among the retrieved samples. A GNN then processes this graph to generate an enriched representation that is prepended to the query sequence, providing additional context for improved event prediction (Fig.\ref{framework}(c), Sec.\ref{sec:graph-fusion-gen}). By integrating retrieval and graph fusion, RAG4DyG effectively incorporates temporal and contextual relevance, surpassing existing methods in both transductive and inductive dynamic graph scenarios.

\subsection{Time- and Context-Aware Retriever}
\label{sec:retriever}
Unlike NLP retrievers, dynamic graph retrieval requires consideration of temporal proximity alongside contextual relevance. To address this, we propose a time- and context-aware retrieval model with two contrastive learning modules. First, we incorporate a time decay mechanism to account for temporal proximity between query and candidate sequences. Second, we use sequence augmentation to capture intrinsic contextual patterns.
%Unlike the retriever in NLP, temporal proximity is important for dynamic graph retrieval in addition to contextual relevance. To this end, we propose a time- and context-aware retrieval model consisting of two contrastive learning modules to jointly consider both aspects. First, we integrate a time decay mechanism to account for the temporal proximity between the query and candidate sequences. Second, we apply sequence augmentation to learn the intrinsic contextual patterns. %To achieve this, we represent the dynamic graph as a series of sequences for ego nodes inspired by SimpleDyG \cite{wu2024feasibility}. We first train the SimpleDyG to learn the dynamic and then fine-tune the Transformer module in SimpleDyG as the encoder for the two modules in the retriever, which are introduced as follows.

\vspace{+0.1cm}\noindent\textbf{Retrieval Annotation.} To facilitate contrastive training, we automatically annotate the samples in the retrieval pool $D$. For each query sequence $x_q$, we annotate its positive sample $x_p^+$ from the pool $D$ based on their contextual similarity. We leave the detailed annotation process in Sec.~\ref{sec:exp-setup}. Specifically, we adopt the sequence model pre-trained in Fig.~\ref{framework}(a) as the encoder and apply mean pooling to obtain sequence representations. Given a query sequence $x_q$ and a candidate sequence $x_p \in D $, we define their contextual similarity as the dot product of their representations:
\begin{equation}
    s(x_q,x_p)=f(x_q)^\top f(x_p),
\end{equation}
where $f(\cdot)$ denotes our encoder.

%In what follows, we introduce the details of the two constrative learning modules.

\vspace{+0.1cm}\noindent\textbf{Time-aware Contrastive Learning.} Temporal information reflects the dynamic changes in historical interactions, which is crucial for dynamic graph modeling. We posit that demonstrations closer in time to the query are more relevant than those further away. Consequently, we utilize a time decay function to account for temporal proximity between the query and candidate sequences, as follows.
\begin{equation}
    \mu (t_q,t_p) = \exp(-\lambda|t_q-t_p|), \label{eq: decay}
\end{equation}
where $t_q$ and $t_p$ represent the last interaction time in the query and candidate sequences\footnote{In the annotated training data, the query time $t_q$ may precede the candidate time $t_p$. However, in the validation and test sets, $t_p$ always precedes $t_q$, preventing leakage from a future time.}, respectively. The hyper-parameter $\lambda$ controls the rate of time decay, determining how quickly the importance of interactions decreases with time. Note that $0<\mu(\cdot,\cdot)\le 1$. By using this time decay function, we assign higher importance to the candidates that are temporally closer to the query.

To effectively capture the temporal dynamics of the graph, we incorporate temporal proximity to reweigh the contextual similarity in the contrastive loss:
\begin{align}
h(x_q,x_p) &= s(x_q,x_p) \mu (t_q,t_p).
\end{align}
Subsequently, we adopt in-batch negative sampling based on the following training objective:
\begin{align}
 \mathcal{L}_{\text{tcl}} &= -\log \frac{\exp (h(x_q, x_p^+))  / \tau}{\sum_{j=1}^{2N}{\mathds{1}}_{j \neq q} \exp (h(x_q, x_j))  / \tau},
\end{align}
where $x_p^+$ denotes the positive sample of $x_q$, $N$ is the batch size, and
 $\tau$ is the temperature parameter. 

\vspace{+0.1cm}\noindent\textbf{Context-aware Contrastive Learning.}
To better capture the inherent contextual pattern, we further adopt context-aware contrastive learning with data augmentations. For each sequence, we apply two types of augmentations: masking and cropping, which are widely used for sequence modeling \cite{devlin2019bert,xie2022contrastive}. The masking operator randomly replaces a portion of the tokens in the sequence with a special masking token. The cropping operator randomly deletes a contiguous subsequence from the original sequence, reducing the sequence length while preserving the temporal order of the interactions. These augmentations help the model learn robust representations and capture the inherent structural information of the sequence by focusing on its different parts. 

We treat two augmented views of the same sequence as positive pairs, and those of different sequences as negative pairs.
%We treat the anchor and augmented sequences as positive pairs and other sequences in the batch as negative samples. 
Given a sequence $x_q$ and its two distinct augmented views $x'_q$ and $x''_q$, the contrastive loss is defined as:
\begin{equation}
 \mathcal{L}_{\text{ccl}} = -\log \frac{\exp (s(x'_q, x''_q) / \tau)}{\sum_{j=1}^{2N}{\mathds{1}}_{j \neq q} \exp s(x'_q, x'_j) / \tau},
% \mathcal{L}_{\text{ccl}} = -\log \frac{\exp (s(x_q^a , x_q^{a'}) / \tau)}{\sum_{j=1}^{2N}{\mathds{1}}_{j \neq q} \exp s(x_q^a, x_j^a) / \tau},
\end{equation}
where $\tau$ is the temperature, $N$ is the batch size and $\mathds{1}$ is an indicator function.

%\subsubsection{Training Objective of Retrieval Model.}
\vspace{+0.1cm}\noindent\textbf{Training and Inference for Retrieval.}
The training objective of our retrieval model is defined as:
\begin{equation}
 \mathcal{L}_{\text{ret}} = \mathcal{L}_{\text{tcl}} + \alpha \mathcal{L}_{\text{ccl}}, \label{eq: retloss}
\end{equation}
where $\alpha$ is a coefficient that balances between the two losses. 
During testing, we utilize the updated sequence model to extract sequence representations and perform demonstration ranking
based on the contextual similarity between the query and candidates.
%compute the dot products between them to perform similar demonstration ranking.

\subsection{Graph Fusion-based Generator}\label{sec:graph-fusion-gen}

After the retrieval process, we obtain the top-\textit{K} demonstrations $R_q = \{(x_k, y_k)\}_{k=1}^{K}$ for the query $x_q$. A straightforward approach is to directly concatenate them with the query sequence and input to a sequence generation model for prediction. However, this can lead to a lengthy context that limits the model's prediction capabilities. More importantly, it neglects the structural patterns among these demonstrations. Thus, we first fuse the demonstrations into a summary graph, process it using a GNN, and then prepend the graph readout from the GNN to the query for subsequent generation.

\vspace{+0.1cm}\noindent\textbf{Graph Fusion.}
 To effectively fuse the demonstrations in $R_q$, we construct a summary graph, whose nodes include all tokens in the retrieved demonstrations, and edges represent the interactions between nodes within each sequence. 
 %For each demonstration tuple $(x_k, y_k)$, the nodes are all the tokens in the tuple, and the edges are the links between the ego node and other tokens in the tuple. 
 Considering that there are common tokens across the retrieved demonstrations (\eg, recurring nodes in multiple demonstrations and special tokens like $[\textit{hist}]$, $[\textit{time1}]$, \etc), we can fuse these demonstrations into a summary graph  $G_{\text{fus}}$.
% $G_{\text{fus}} = (\mathcal{V},\mathcal{E})$, where $\mathcal{V}$ denotes the node set in the top-\textit{K} demonstrations, and $\mathcal{E}$ denotes the edges.  
 We then employ a graph convolutional network (GCN) to capture the structural and contextual information within the fused graph and apply a mean-pooling readout to obtain a representation vector for the graph. The vector is subsequently concatenated with the query sequence representation, as follows.
\begin{align}
%x_i &= \langle |\textit{hist}| \rangle, v_i, \langle |\textit{time1}|\rangle,  v_i^1, \dots v_i^{w-1}, \langle |\textit{endofhist}|\rangle \\
e_{\text{fus}} &= \mathtt{MeanPooling}(\text{GCN}(G_{\text{fus}})), \label{eq:fus}\\
 %x_q' &= \langle |\textit{hist}| \rangle, e_{\text{fus}}, v_i, \langle |\textit{time1}|\rangle,  v_i^1, \dots v_i^{w-1}, \langle |\textit{eohist}|\rangle. 
  \tilde{x}_q &= [e_{\text{fus}} \parallel x_q], \label{eq:fus-concat} 
  %x_q' &= e_{\text{fus}},\langle |\textit{hist}|\rangle,  v_q, \langle |\textit{time1}|\rangle,  S_q^1, \dots \langle |\textit{timeT}|\rangle, S_q^{T}, \langle |\textit{eohist}|\rangle, \\
  %S_q^t &= \langle v_q^{t,1}, v_q^{t,2} \dots v_q^{t,|S_q^t|} \rangle.
\end{align}
where $e_{\text{fus}}$ is the fused graph representation, and $\tilde{x}_q$ is the retrieval-augmented sequence. The augmented sequence is fed into the sequence model, which generates future interactions. 

\vspace{+0.1cm}\noindent\textbf{Training and Inference.} We adopt the same sequence model with the same training objective \cite{wu2024feasibility} as in Fig.~\ref{framework}(a). During training, we freeze the parameters of the sequence model, except for the output layer which is updated along with the GCN parameters used for graph fusion.
During testing, we first apply the retriever model to retrieve top-\textit{K} demonstrations for each query as introduced in Sec.~\ref{sec:retriever}. Then we perform graph fusion on these demonstrations and concatenate the fused graph representation with the query sequence as illustrated in Eq.~\eqref{eq:fus} and \eqref{eq:fus-concat}. The concatenated sequence is subsequently fed into the trained model for link prediction.

\section{Experiment}
In this section, we empirically evaluate the proposed model RAG4DyG compared to state-of-the-art methods and conduct a detailed analysis of the performance.

\subsection{Experimental Setup}
\label{sec:exp-setup}
\subsubsection{Datasets.} We evaluate the performance of the proposed model on six datasets from different domains, including a communication network UCI \cite{panzarasa2009patterns}, a citation network Hepth \cite{leskovec2005graphs}, a multi-turn task-oriented conversation dataset MMConv \cite{liao2021mmconv}, a behavioral interaction network %tgbl-wiki-v2.
Wikipedia \cite{huang2024temporal}, an email network Enron \cite{zhang2024dtgb}, and a hyperlink network Reddit \cite{kumar2018community}. We summarize the statistics of these datasets in Table~\ref{dataset}. We follow the preprocessing steps of SimpleDyG to map the dynamic graphs into sequences with special tokens \cite{wu2024feasibility}.  
%For UCI, Hepth and MMConv datasets, we follow the same preprocessing as SimpleDyG to map the dynamic graphs into sequences with special tokens \cite{wu2024feasibility}. It is important to note that the ML-10M dataset was not used because it is not suitable for RAG; we observed no performance enhancement even when the query was augmented with ground-truth demonstrations. 
Notably, the Hepth and Reddit datasets exhibit an inductive nature, as they contain previously unseen nodes with no historical interactions. The details of the Wikipedia, Enron and Reddit datasets are provided below, while information on  the remaining datasets can be found in SimpleDyG \cite{wu2024feasibility}. 

 %\textcolor{blue}{\textbf{UCI} \cite{panzarasa2009patterns}: This dataset captures a social network where edges denote messages exchanged between users. To ensure temporal alignment, we follow prior work \cite{sankar2020dysat} and divide the data into 13 time steps.}

%\textcolor{blue}{\textbf{Hepth} \cite{leskovec2005graphs}: This dataset represents a citation network within the high-energy physics theory community. Temporal alignment is achieved by extracting 24 months of data and dividing it into 12 time steps. The initial node features are derived from the abstracts of the papers, processed using the word2vec model \cite{mikolov2013distributed}. }

%\textcolor{blue}{\textbf{MMConv} \cite{liao2021mmconv}: The dataset comprises a multi-turn task-oriented dialogue system designed to assist users in discovering locations across five distinct domains. Utilizing its comprehensive annotations, the dialogue interactions are modeled as a dynamic graph, a commonly adopted approach in task-oriented dialogue systems. Temporal alignment is achieved by dividing the data into 16 time steps, with each time step representing a specific turn in the conversation.}

\begin{itemize}[leftmargin=*]
\item \textbf{Wikipedia} \cite{huang2024temporal}:   
This dataset captures the co-editing activity on Wikipedia pages over one month. It is a bipartite interaction network in which editors and wiki pages serve as nodes. Each edge corresponds to an interaction where a user edits a page at a specific timestamp. To facilitate temporal sequence alignment, the dataset is divided into 16 time steps based on the timestamps of the interactions.

\item \textbf{Enron} \cite{zhang2024dtgb}: This dataset represents the  
email communications among employees of Enron Corporation over three years (1999--2002). Nodes represent employees, while edges correspond to emails exchanged between them ordered by the sending timestamps of the emails. For temporal sequence alignment, we split the dataset into 17 time steps based on the timestamps.

\item \textbf{Reddit} \cite{kumar2018community}: This dataset represents a subreddit-to-subreddit hyperlink network, derived from timestamped posts containing hyperlinks between subreddits. %, with provided subreddit embeddings. 
We focus on hyperlink data within the body of posts, covering the period from 2016 to 2017. The dataset is divided into 12 time steps for temporal sequence alignment based on the post timestamps.
\end{itemize}

%\end{itemize}
%To facilitate retrieval model training, we regard the samples in the training dataset as our retrieval source and automatically annotate them as introduced in Appendices \ref{app-anno}. %We annotate demonstrations based on the Jaccard of future interaction of each sample, which is the interactions at the last timestep in training data (the sequence with red rectangles in the retrieval pool of Figure \ref{framework}). To control the quality of annotated data, we set a threshold of 0.8 to select high-related demonstrations. 
%The statistical information of the three datasets with the number of training samples for retrieval is shown in Table \ref{dataset}.
%\vspace{-0.2cm}
\begin{table}[t]
  \caption{Dataset statistics.}
  \vspace{-0.2cm}
    \centering
    \small
    \addtolength{\tabcolsep}{-2.5pt}
    \renewcommand*{\arraystretch}{1.3}
    \begin{tabular}{ccccccc}
    \toprule
    Dataset &  UCI  &Hepth & MMConv &Wikipedia & Enron & Reddit\\
    \midrule
    Domain & Social     &Citation &Conversation & Behavior &Social &Hyperlink\\
    {\#} Nodes &1,781     & 4,737  &7,415 &9,227 &42,711 &11,901 \\
    {\#} Edges &16,743      &14,831  &91,986 &157,474 & 797,907 &62,919   \\ 
%    {\#} Retrieval &9,578      &8,250  &10,762 \\     
  \bottomrule
\end{tabular}
  \label{dataset}
  \vspace{-0.2cm}
\end{table}
%\vspace{-0.2cm}

\begin{table*}[ht]
\centering
\caption{Performance comparison for dynamic link prediction with mean and standard deviation across 10 runs. Best results are \textbf{bolded}; runners-up are \underline{underlined}. * indicates that our model significantly outperforms the best baseline based on the two-tail $t$-test ($p< 0.05$).}
\vspace{-0.2cm}
\label{result}
\small
\addtolength{\tabcolsep}{-2.7pt}
\renewcommand*{\arraystretch}{1.3}
\begin{tabular}{c|c|ccccccccccc}
\hline
Datasets & Models & DySAT & EvolveGCN & DyRep & JODIE  & TGAT & TGN & TREND & GraphMixer & IDOL& SimpleDyG& RAG4DyG \\ \hline

\multirow{3}{*}{UCI}     
& Recall@5       
& 0.009\scriptsize{±0.003} 
& 0.072\scriptsize{±0.046} 
& 0.009\scriptsize{±0.008}
& 0.018\scriptsize{±0.019}
& 0.022\scriptsize{±0.004}
& 0.014\scriptsize{±0.010} 
& 0.083\scriptsize{±0.015}
& 0.097\scriptsize{±0.019}
& 0.093\scriptsize{±0.029}
& \underline{0.109}\scriptsize{±0.014}
& \textbf{0.111}\scriptsize{±0.013} 
\\ 
& NDCG@5          
& 0.010\scriptsize{±0.003} 
& 0.064\scriptsize{±0.045}
& 0.011\scriptsize{±0.018}
& 0.022\scriptsize{±0.023}
& 0.061\scriptsize{±0.007}
& 0.041\scriptsize{±0.017} 
& 0.067\scriptsize{±0.010}
& \underline{0.104}\scriptsize{±0.013}
& 0.075\scriptsize{±0.022} 
& \underline{0.104}\scriptsize{±0.010}  
& \textbf{0.122*}\scriptsize{±0.014}
\\ 
& Jaccard              
& 0.010\scriptsize{±0.001}  
& 0.032\scriptsize{±0.026}
& 0.010\scriptsize{±0.005}
& 0.012\scriptsize{±0.009}
& 0.020\scriptsize{±0.002}
& 0.011\scriptsize{±0.003} 
& 0.039\scriptsize{±0.020} 
& 0.042\scriptsize{±0.005}  
& 0.014\scriptsize{±0.002} 
& \underline{0.092}\scriptsize{±0.014} 
& \textbf{0.097}\scriptsize{±0.010} 
\\ \hline

\multirow{3}{*}{Hepth}     
& Recall@5       
& 0.008\scriptsize{±0.004}
& 0.008\scriptsize{±0.002}
& 0.009\scriptsize{±0.006}
& 0.010\scriptsize{±0.008}
& 0.011\scriptsize{±0.007} 
& 0.011\scriptsize{±0.006}
& 0.010\scriptsize{±0.008}
& 0.009\scriptsize{±0.002}
& 0.007\scriptsize{±0.002}
& \underline{0.013}\scriptsize{±0.006}
& \textbf{0.019*}\scriptsize{±0.002} 
\\ 
& NDCG@5          
& 0.007\scriptsize{±0.002}  
& 0.009\scriptsize{±0.004} 
& 0.031\scriptsize{±0.024} 
& 0.031\scriptsize{±0.021} 
& 0.034\scriptsize{±0.023}  
& 0.030\scriptsize{±0.012}  
& 0.031\scriptsize{±0.003} 
& 0.011\scriptsize{±0.008}
& 0.011\scriptsize{±0.003}
& \underline{0.035}\scriptsize{±0.014}    
& \textbf{0.045*}\scriptsize{±0.003} 
\\ 
& Jaccard              
& 0.005\scriptsize{±0.001}  
& 0.007\scriptsize{±0.002} 
& 0.010\scriptsize{±0.006} 
& 0.011\scriptsize{±0.008}
& 0.011\scriptsize{±0.006} 
& 0.008\scriptsize{±0.001} 
& 0.010\scriptsize{±0.002} 
& 0.010\scriptsize{±0.003} 
& 0.006\scriptsize{±0.001}
& \underline{0.013}\scriptsize{±0.006} 
& \textbf{0.019*}\scriptsize{±0.002}
\\ \hline

\multirow{3}{*}{MMConv}     
& Recall@5       
& 0.108\scriptsize{±0.089} 
& 0.050\scriptsize{±0.015}
& 0.156\scriptsize{±0.054} 
& 0.052\scriptsize{±0.039}
& 0.118\scriptsize{±0.004}
& 0.085\scriptsize{±0.050} 
& 0.134\scriptsize{±0.030} 
& \textbf{0.206}\scriptsize{±0.001} 
& 0.169\scriptsize{±0.006}
& 0.170\scriptsize{±0.010}
& \underline{0.194}\scriptsize{±0.005} 
\\ 
& NDCG@5          
& 0.102\scriptsize{±0.085} 
& 0.051\scriptsize{±0.021} 
& 0.140\scriptsize{±0.057} 
& 0.041\scriptsize{±0.016}  
& 0.089\scriptsize{±0.033}  
& 0.096\scriptsize{±0.068} 
& 0.116\scriptsize{±0.020} 
& 0.172\scriptsize{±0.029}  
& 0.115\scriptsize{±0.039}
& \underline{0.184}\scriptsize{±0.012} 
& \textbf{0.208*}\scriptsize{±0.005}
\\ 
& Jaccard              
& 0.095\scriptsize{±0.080} 
& 0.032\scriptsize{±0.017} 
& 0.067\scriptsize{±0.025}
& 0.032\scriptsize{±0.022}
& 0.058\scriptsize{±0.021} 
& 0.066\scriptsize{±0.038}  
& 0.060\scriptsize{±0.018}  
& 0.085\scriptsize{±0.016}  
& 0.015\scriptsize{±0.002}
& \underline{0.169}\scriptsize{±0.010} 
& \textbf{0.194*}\scriptsize{±0.005} 
\\ \hline

\multirow{3}{*}{Wikipedia}     
& Recall@5       
& 0.003\scriptsize{±0.005}  
& 0.012\scriptsize{±0.01} 
& 0.003\scriptsize{±0.002} 
& 0.017\scriptsize{±0.005} 
& 0.006\scriptsize{±0.004}
& 0.016\scriptsize{±0.018}  
& 0.022\scriptsize{±0.012}
& 0.010\scriptsize{±0.007}
& 0.022\scriptsize{±0.008}
&\underline{0.356}\scriptsize{±0.016}
& \textbf{0.369*}\scriptsize{±0.006}
\\ 
& NDCG@5          
& 0.002\scriptsize{±0.003}  
& 0.008\scriptsize{±0.007} 
& 0.002\scriptsize{±0.002} 
& 0.015\scriptsize{±0.003} 
& 0.005\scriptsize{±0.005}
& 0.015\scriptsize{±0.022} 
& 0.016\scriptsize{±0.018}
& 0.007\scriptsize{±0.006}
& 0.015\scriptsize{±0.005}
& \textbf{0.398}\scriptsize{±0.03}
& \underline{0.389}\scriptsize{±0.008} 
\\ 
& Jaccard              
& 0.001\scriptsize{±0.001}  
& 0.004\scriptsize{±0.004} 
& 0.001\scriptsize{±0.001} 
& 0.007\scriptsize{±0.002} 
& 0.002\scriptsize{±0.002}
& 0.007\scriptsize{±0.009} 
& 0.007\scriptsize{±0.021} 
& 0.004\scriptsize{±0.002}
& 0.004\scriptsize{±0.001}
& \underline{0.320}\scriptsize{±0.027}
& \textbf{0.328}\scriptsize{±0.007} 
\\ \hline

\multirow{3}{*}{Enron}     
& Recall@5       
& 0.002\scriptsize{±0.004} 
& 0.004\scriptsize{±0.011}
& 0.021\scriptsize{±0.001} 
& 0.005\scriptsize{±0.005} 
& 0.020\scriptsize{±0.002} 
& 0.001\scriptsize{±0.001} 
& 0.023\scriptsize{±0.003}  
& 0.021\scriptsize{±0.002} 
& 0.024\scriptsize{±0.014} 
& \underline{0.094}\scriptsize{±0.005} 
& \textbf{0.100*}\scriptsize{±0.003} 
\\ 
& NDCG@5          
& 0.001\scriptsize{±0.002}  
& 0.007\scriptsize{±0.020}
& 0.036\scriptsize{±0.002} 
& 0.061\scriptsize{±0.039}  
& 0.036\scriptsize{±0.001} 
& 0.003\scriptsize{±0.001} 
& 0.027\scriptsize{±0.001}  
& 0.037\scriptsize{±0.001} 
& 0.025\scriptsize{±0.011} 
& \underline{0.114}\scriptsize{±0.005}
& \textbf{0.119*}\scriptsize{±0.004} 
\\ 
& Jaccard              
& 0.001\scriptsize{±0.001}  
& 0.003\scriptsize{±0.009} 
& 0.019\scriptsize{±0.001} 
& 0.011\scriptsize{±0.007}  
& 0.020\scriptsize{±0.001} 
& 0.001\scriptsize{±0.001} 
& 0.012\scriptsize{±0.001}  
& 0.020\scriptsize{±0.002} 
& 0.008\scriptsize{±0.003}
& \underline{0.068}\scriptsize{±0.003}
& \textbf{0.071*}\scriptsize{±0.002} 

\\  \hline
\multirow{3}{*}{Reddit}     
& Recall@5       
& 0.001\scriptsize{±0.002}
& 0.006\scriptsize{±0.002}
& 0.019\scriptsize{±0.004}
& 0.013\scriptsize{±0.003}
& 0.001\scriptsize{±0.001}
& 0.001\scriptsize{±0.001}
& 0.002\scriptsize{±0.003}
& 0.001\scriptsize{±0.001}
& 0.003\scriptsize{±0.002}
& \underline{0.101}\scriptsize{±0.019} 
& \textbf{0.119*}\scriptsize{±0.006} 

\\
& NDCG@5       
& 0.001\scriptsize{±0.002}
& 0.012\scriptsize{±0.003}
& 0.020\scriptsize{±0.004}
& 0.015\scriptsize{±0.002}
& 0.001\scriptsize{±0.001}
& 0.002\scriptsize{±0.001}
& 0.003\scriptsize{±0.002}
& 0.003\scriptsize{±0.001}
& 0.005\scriptsize{±0.003}
& \underline{0.134}\scriptsize{±0.012} 
& \textbf{0.143}\scriptsize{±0.005} 

\\
& Jaccard      
& 0.001\scriptsize{±0.001}
& 0.003\scriptsize{±0.001}
& 0.013\scriptsize{±0.004}
& 0.007\scriptsize{±0.002}
& 0.001\scriptsize{±0.001}
& 0.001\scriptsize{±0.002}
& 0.001\scriptsize{±0.001}
& 0.002\scriptsize{±0.001}
& 0.002\scriptsize{±0.001}
& \underline{0.088}\scriptsize{±0.012} 
& \textbf{0.096}\scriptsize{±0.003} 
\\
\hline

\end{tabular}
\vspace{-0.2cm}
\end{table*}

\subsubsection{Baselines.} We compare our model RAG4DyG with the state-of-the-art dynamic graph models, which include (1) discrete-time approaches: DySAT \cite{sankar2020dysat} and EvolveGCN \cite{pareja2020evolvegcn}; (2) continuous-time approaches: DyRep \cite{trivedi2019dyrep}, JODIE \cite{kumar2019predicting}, TGAT \cite{xu2020inductive}, TGN \cite{rossi2020temporal}, TREND \cite{ wen2022trend}, GraphMixer \cite{cong2022we}, IDOL \cite{zhu2024topology} and SimpleDyG \cite{wu2024feasibility}. %The details of the baselines can be found in Appendix B. 

\begin{itemize}[leftmargin=*]
\item \textbf{DySAT} \cite{sankar2020dysat} utilizes self-attention mechanisms to capture both structural and temporal patterns in dynamic graphs through discrete-time snapshots.

\item \textbf{EvolveGCN} \cite{pareja2020evolvegcn}  leverages recurrent neural networks to model the evolution of the parameters of a graph convolutional network over discrete time steps.

\item \textbf{DyRep} \cite{trivedi2019dyrep} models dynamic graphs in continuous time by incorporating both temporal point processes and structural dynamics to capture interactions and node dynamics.

\item \textbf{JODIE} \cite{kumar2019predicting}  focuses on user and item embedding trajectories over continuous time, predicting future interactions by modeling user and item embeddings jointly.

\item \textbf{TGAT} \cite{xu2020inductive}  employs temporal graph attention layers and time encoding to capture temporal dependencies and structural information for dynamic graphs.

\item \textbf{TGN} \cite{rossi2020temporal} combines GNNs with memory modules to maintain node states over continuous time, effectively learning from dynamic interactions.

\item \textbf{TREND} \cite{wen2022trend} integrates temporal dependencies based on the Hawkes process and GNNs to learn the dynamics of graphs.

\item \textbf{GraphMixer} \cite{cong2022we} 
introduces a novel architecture that leverages MLP-mixer to learn link-encoder and node encoder for evolving graphs in continuous time.

\item \textbf{IDOL} \cite{zhu2024topology} is a contrastive learning-based model tailored for dynamic graph representation learning. It utilizes a Personalized PageRank-based algorithm to incrementally update the node embedding and adopt a topology-monitorable sampling method to generate contrastive pairs for efficient training.

\item \textbf{SimpleDyG} \cite{wu2024feasibility} reformulates the dynamic graph modeling as a sequence modeling task and mapped the dynamic interactions of target nodes as sequences with specially designed tokens. It simplifies dynamic graph modeling without complex architectural changes to effectively capture temporal dynamics.
 \end{itemize}

%\vspace{-0.4cm}
\subsubsection{Implementation Details.} Following the method outlined in \cite{cong2022we,wu2024feasibility}, we represent the dynamic graph as an undirected graph. We split all datasets into training, validation, and test sets based on temporal sequence same as SimpleDyG \cite{wu2024feasibility}. Given ${T}$ timesteps in each dataset, the data at the final timestep ${T}$ is used as the testing set, the data at ${T-1}$ is served as the validation set, and the remaining data from earlier timesteps is used for training. All training data including the retrieval pool for the retriever and generator is drawn from this training data split. For retrieval augmented generation model training, we first train SimpleDyG without augmentation using the finetuned parameters. Then we fix the parameters of SimipleDyG except for the last linear layer and fine-tune them with the GCN model. The number of GCN layers in the generator model is 1 for all datasets. We repeat each experiment 10 times and report the average results along with the standard deviation. The number of demonstrations is 7 for all datasets. 

To facilitate retrieval model training, we regard the samples in the training dataset as our retrieval pool $D=\{(x_i, y_i)\}_{i=1}^{M}$ where each pair $(x_i, y_i)$ represents the historical sequence and its corresponding target sequence. Specifically, $x_i$ is the input sequence before the last time step and $y_i$ is the output sequence at the last time step. We annotate demonstrations based on the Jaccard similarity between the output sequences among all the pairs in $D$. The Jaccard similarity of two output sequences, $y_i$ and $y_j$, is given by
\begin{equation}
    r(y_i,y_j) = \frac{|y_i \cap y_j|}{|y_i \cup y_j|}.
\end{equation}
To control the quality of annotated data, we set a threshold of 0.8 to select highly similar demonstrations for each sample. These filtered annotations are then used to train the retriever model. The number of training samples for the UCI, Hepth, MMConv, Wikipedia, Enron and Reddit datasets are 9\,578, 8\,250, 10\,762, 162\,408, 2\,510\,666 and 185\,764, respectively. 
 
We ran all the experiments on a Nvidia L40 GPU and tuned the hyper-parameters for all the methods based on the validation set. For all the baselines, we tuned the models based on the hyper-parameters reported in their papers. For our RAG4DyG method, we set the number of layers, heads and dimensions of hidden states for the backbone SimpleDyG to $(6, 8, 768)$, $(12, 2, 256)$, $(2, 2, 256)$, $(2, 6, 768)$, $(2, 6, 768)$, and $(2, 8, 512)$ across the six datasets. The time decay rate $\lambda$ for the retrieval model in Eq.~\eqref{eq: decay} was tuned according to the time granularity of different datasets, with days for the UCI dataset, months for the Hepth dataset, turns for the MMConv dataset, and hours for other datasets. We explored a range of values $\lambda = \{10^{-4}, 10^{-3}, 10^{-2}, 10^{-1}, 1, 10, 100\}$, ultimately selecting $\lambda = 10^{-4}$ for UCI, $\lambda = 0.1$ for Hepth, $\lambda = 10$ for MMConv and Enron, and $\lambda = 1$ for Wikipedia and Reddit datasets. The coefficient $\alpha$ in the loss function in Eq.~\eqref{eq: retloss} was tuned across $\{0.2, 0.4, 0.6, 0.8, 1\}$, resulting in final values of $\alpha = 1$ for UCI and MMConv, $\alpha = 0.4$ for Hepth, and $\alpha = 0.2$ for Wikipedia, Enron and Reddit. Additional parameter settings for the three datasets are as follows: the temperature $\tau$ in the two contrastive learning tasks for all datasets was set to  $\tau = 0.1$, the batch size of the retriever model for all datasets was set to $N = 128$, and the masking and cropping portions in context-aware contrastive learning were set to 0.8, 0.8, 0.8, 0.6, 0.6, 0.2 and 0.4, 0.6, 0.6, 0.8, 0.8, 0.8 across the six datasets, respectively. 

\subsubsection{Evaluation Metrics.} Inspired by SimpleDyG \cite{wu2024feasibility}, we assess the performance of our approach and baselines using three key metrics: Recall@5, NDCG@5, and Jaccard \cite{wu2024feasibility}. Recall@5 and NDCG@5 are commonly employed in ranking tasks to evaluate the quality of top-ranked predictions \cite{wang2019neural}. Specifically, Recall@5 measures the proportion of relevant nodes that appear among the top five predictions, while NDCG@5 considers the ranking positions of the relevant nodes to provide a more nuanced assessment of ranking quality. Additionally, the Jaccard index \cite{jaccard1901etude} quantifies the similarity between the predicted and ground truth sequences by calculating the ratio of their intersection to their union.

\begin{figure*}[t] 
\centering
    \centering
    \includegraphics[scale=1]{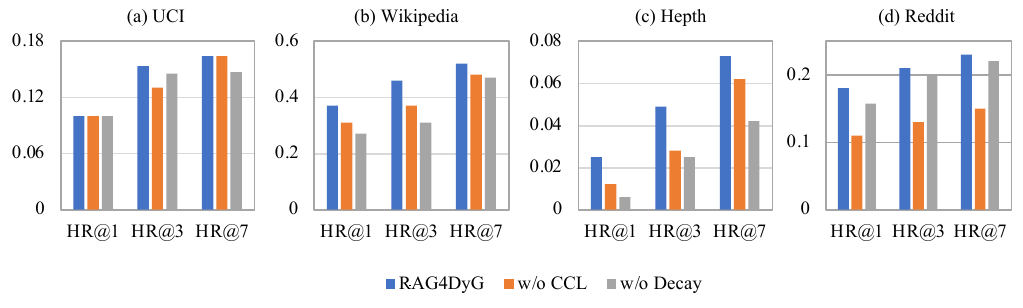}
    \vspace{-0.4cm}
\caption{Ablation study for retrieval results.}
\label{ablation-re}
\end{figure*}

\begin{figure*}[!htp] 
\centering
    \centering
    \includegraphics[scale=1]{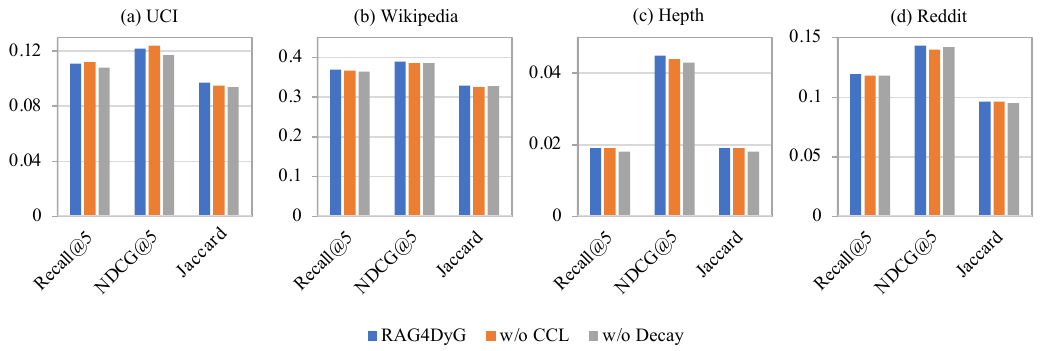}
    \vspace{-0.3cm}
    \caption{Ablation study for link prediction results.}
    \label{ablation-gen}
   % \vspace{-0.2cm}
\end{figure*}

\subsection{Performance Comparison}

We assess the performance of RAG4DyG on the dynamic link prediction task, with the results benchmarked against state-of-the-art baselines, as shown in Table \ref{result}. Our analysis reveals the following key observations.

First of all, the proposed RAG4DyG generally outperforms all baselines across different datasets under the three metrics. 
In particular, compared to SimpleDyG, which is also our backbone, RAG4DyG consistently shows superior performance, highlighting the effectiveness of our retrieval-augmented generation framework. 
Note that GraphMixer performs slightly better in Recall@5 on the MMConv dataset, but its significantly lower performance in NDCG@5 and Jaccard indicates that its predictions are not ranked optimally or maintaining the overall set integrity compared to RAG4DyG. This indicates that RAG4DyG can better model the temporal and contextual relationships due to the specific design of the retriever and generator. Generally speaking, the performance of SimpleDyG and RAG4DyG which reformulate the dynamic graph link prediction as a sequence generation task show promising performance compared with node pair ranking-based baselines, especially on the Wikipedia dataset which contains a higher frequency of repeated interaction behaviors. This characteristic makes sequence-based models more effective, as they can leverage the temporal consistency and recurrent patterns in the data to better capture the underlying dynamics of the graph.
%This discrepancy could be attributed to GraphMixer's architecture, which relies on a link encoder and node encoder to summarize information from temporal links and neighbors. While effective for predicting possible links, this approach does not excel at capturing inherent sequence properties and thus is inferior in predicting the full ranking or set. %. 
%RAG4DyG, on the other hand, can better model the temporal and contextual relationships due to the specific design of the retriever and generator.

Second, RAG4DyG exhibits significant advantages in inductive scenarios such as the Hepth and Reddit datasets. This setting is particularly challenging because it involves nodes not seen during training, requiring the model to generalize to entirely new structures and relationships. 
RAG4DyG's success is attributed to its retrieval-augmented mechanism, which enhances the model's ability to generalize by providing rich contextual information relevant to the new, unseen nodes. Unlike models that rely solely on the immediate neighborhood or predefined structures, RAG4DyG dynamically adapts to the new nodes, ensuring that the predictions are guided by the most relevant and similar historical data.

\subsection{Model Analysis}
We analyze the behavior of our model RAG4DyG in several aspects, including an ablation study, an investigation of the effectiveness of different retrieval methods, and an analysis of parameter sensitivity.

\begin{table*}[htp]
    %\vspace{-1mm}
    \caption{Retrieval performance of various retrieval methods. }
    \vspace{-0.2cm}
	\centering
	\small
	%\addtolength{\tabcolsep}{-3pt}
	\addtolength{\tabcolsep}{3pt}
	\renewcommand*{\arraystretch}{1.3}
    %   \vspace{-0.1cm}
    %\setlength{\tabcolsep}{2pt}
    %\resizebox{0.47\textwidth}{!}{
    
    %\begin{tabular}{c|ccc|ccc|ccc|ccc|ccc|ccc}
    \begin{tabular}{c|ccc|ccc|ccc|ccc}
	\toprule
	\multicolumn{1}{c|}{\multirow{2}{*}{Method}}
        & \mc{3}{c|}{UCI}  
        & \mc{3}{c|}{Wikipedia} 
        & \mc{3}{c|}{Hepth} 
        %& \mc{3}{c|}{MMConv} 
        
        %& \mc{3}{c|}{Enron} 
        & \mc{3}{c}{Reddit}\\
	%\cmidrule{2-19}
	\cmidrule{2-13}
	& {HR@1} & {HR@3} & {HR@7}
        & {HR@1} & {HR@3} & {HR@7}    
        & {HR@1} & {HR@3} & {HR@7}
        %& {HR@1} & {HR@3} & {HR@7}
        %& {HR@1} & {HR@3} & {HR@7} 
        & {HR@1} & {HR@3} & {HR@7} 
        \\		
	\midrule		
	\textit{BM25}  
        & \textbf{0.100}  & 0.136  & \textbf{0.200}
        & \textbf{0.369}  & 0.405 & 0.488
        & -   & -  & -
        %& 0.025 &\textbf{0.233} & 0.356
        
        %& 0.090  & \textbf{0.126} & \textbf{0.168}
        & -   & -  & -
        \\ 
	\textit{Jaccard}  
        & \textbf{0.100}  &0.109 & 0.146
        & \textbf{0.369}  & 0.445 & 0.430
        & -   & - & -
        %& 0.007 & 0.213 & 0.342
        
        %& \textbf{0.092}  & 0.119 & 0.158
        & -   & -  & -
        \\        
        \midrule
        \textit{RAG4DyG}  
        & \textbf{0.100} & \textbf{0.155}  & 0.164
        & \textbf{0.369}  &\textbf{0.455} & \textbf{0.523}
        & \textbf{0.025} & \textbf{0.049} &\textbf{0.073}
        %& \textbf{0.034} & 0.168 & 0.273
        
       % & 0.086  &0.115 & 0.160
        & \textbf{0.180}   & \textbf{0.218}  & \textbf{0.228}
        \\ 
		\bottomrule
  \multicolumn{13}{l}{}\\[-5mm]
    \multicolumn{13}{l}{\scriptsize ``-'' denotes that the method is unable to perform retrieval. The reason is explained in the corresponding description of this table in Sec.~5.3. %that the testing nodes in Hepth are all new nodes without historical interactions. We address this by adopting the initial text embeddings of new nodes as the query representation. However, this does not work for the baselines since they rely on the appearance frequency or overlap of the new nodes and candidate sequences.
    }
	\end{tabular}
%}
 \label{re}
\end{table*}

\begin{table*}[h!]
    \centering
    \small
    \addtolength{\tabcolsep}{-3pt}
    %\vspace{-0.1cm}
    \caption{Generative performance of various retrieval methods.}
    \vspace{-0.2cm}
    \renewcommand*{\arraystretch}{1.3}
    %\begin{tabular}{@{}c|ccc|ccc|ccc|ccc|ccc|ccc@{}}
    %\begin{tabular}{@{}c|ccc|ccc|ccc|ccc|ccc@{}}
    \begin{tabular}{c|ccc|ccc|ccc|ccc}
    
    \toprule
    \mc{1}{c|}{\multirow{2}{*}{Method}}
    & \mc{3}{c|}{UCI} 
    & \mc{3}{c|}{Wikipedia} 
    & \mc{3}{c|}{Hepth} 
    %& \mc{3}{c|}{MMConv} 
    
        %& \mc{3}{c|}{Enron} 
        &\mc{3}{c}{Reddit} \\ \cmidrule{2-13}
		
	& {Recall@5} & {NDCG@5} & {Jaccard} & {Recall@5} & {NDCG@5}  & {Jaccard} & {Recall@5} & {NDCG@5} & {Jaccard} %& {R@5} & {NDCG}& {Jaccard}  
    %& {R@5}& {NDCG} & {Jaccard}  
    & {Recall@5} & {NDCG@5} & {Jaccard}
        \\	
		\midrule		
		
		\textit{BM25}  
           &\textbf{0.111}\scriptsize{±0.007} 
           &0.121\scriptsize{±0.009}    
           &0.093\scriptsize{±0.004}

            &0.368\scriptsize{±0.01} 
           & \textbf{0.389}\scriptsize{±0.012} 
           &0.325\scriptsize{±0.01}

            & -   & - &-
            
          % &  0.193%\scriptsize{±0.003}
          % & 0.207%\scriptsize{±0.003}
          % &0.192%\scriptsize{±0.003}  

       % & \textbf{0.105}%\scriptsize{±0.003} 
        %& \textbf{0.124}%\scriptsize{±0.004} 
        %& \textbf{0.075}%\scriptsize{±0.002} 
        & -   & - &-
           
           \\ 
        \textit{Jaccard}
           & 0.104\scriptsize{±0.009} 
           & 0.113\scriptsize{±0.011}    
           & 0.088\scriptsize{±0.010} 

        &0.368\scriptsize{±0.013} 
        & 0.388\scriptsize{±0.014} 
         &0.321\scriptsize{±0.011} 
         
            & -  &- &-
            
         %& 0.193%\scriptsize{±0.004}  
         %& 0.207%\scriptsize{±0.005}
         %&0.192%\scriptsize{±0.004}

        % &0.102%\scriptsize{±0.004} 
       % & 0.122%\scriptsize{±0.004}         
       %  &0.073%\scriptsize{±0.002} 
         & -   & - &-
         \\ 
        \textit{RAG4DyG}  
            & \textbf{0.111}\scriptsize{±0.013} 
            & \textbf{0.122}\scriptsize{±0.014}    
            & \textbf{0.097}\scriptsize{±0.010} 
            
            &\textbf{0.369}\scriptsize{±0.006} 
            &\textbf{0.389}\scriptsize{±0.008} 
            &\textbf{0.328}\scriptsize{±0.007} 
            
            &\textbf{0.019}\scriptsize{±0.002}    
            & \textbf{0.045}\scriptsize{±0.003}   
            & \textbf{0.019}\scriptsize{±0.002}
            
            %&\textbf{0.194}%\scriptsize{±0.005}   
           % & \textbf{0.208}%\scriptsize{±0.005}   
           % & \textbf{0.194}%\scriptsize{±0.005} 

          %  &0.100%\scriptsize{±0.003}  
          %  &0.119%\scriptsize{±0.004} 
          %  &0.071%\scriptsize{±0.002} 
       & \textbf{0.119}\scriptsize{±0.006} 
       & \textbf{0.143}\scriptsize{±0.005} 
       & \textbf{0.096}\scriptsize{±0.003} 
       
       \\         
         \midrule

         \textit{GroundTruth}  
            & 0.121\scriptsize{±0.010}  
            & 0.129\scriptsize{±0.010}  
            & 0.107\scriptsize{±0.012}

            &0.390\scriptsize{±0.008} 
            & 0.400\scriptsize{±0.007}
            &0.340\scriptsize{±0.006}

           & 0.028\scriptsize{±0.004}   
           & 0.062\scriptsize{±0.007}
           &0.028\scriptsize{±0.004}
           
           % & 0.210%\scriptsize{±0.008}    
           % & 0.224%\scriptsize{±0.008}    
           % & 0.208%\scriptsize{±0.008}  

            %&0.116%\scriptsize{±0.002} 
            %& 0.136%\scriptsize{±0.002}
            %&0.087%\scriptsize{±0.002}  
            
            &0.121\scriptsize{±0.008} 
            &0.145\scriptsize{±0.008} 
            &0.099\scriptsize{±0.005} 
        \\
		\bottomrule
    \multicolumn{13}{l}{}\\[-5mm]
    \multicolumn{13}{l}{\scriptsize \ \ \ See the note in Table~\ref{re} for the explanation of ``-''.} %Here R@5 represents Recall@5 and NDCG refers to NDCG@5.}
	\end{tabular}
    
 %\label{app table: re for gen}
 \label{re for gen}
 \vspace{-0.2cm}
\end{table*}

\subsubsection{Ablation Study.}

To evaluate the effectiveness of different modules in the retrieval model, we compare RAG4DyG with two variants \textit{w/o CCL} and \textit{w/o Decay} which exclude the context-aware contrastive learning and time decay component in the retrieval model. We evaluate the performance for both retrieval and link prediction tasks. We use HR@\textit{k} (Hit Ratio@\textit{k}) metrics for the retrieval model, measuring the proportion of cases where at least one of the top-\textit{k} retrieved items is relevant. As shown in Fig.~\ref{ablation-re} and \ref{ablation-gen}, the full model outperforms the two variants, underscoring the benefits of incorporating context-aware contrastive learning and time decay modulation. Notably, the \textit{w/o Decay} variant exhibits the worst performance across both tasks, emphasizing the critical role of time decay in capturing temporal relevance and accurately modeling the evolving dynamics of the graph.

\subsubsection{Effect of Different Retrieval Methods.}

To further investigate the effectiveness of the retrieval model, we compare our model with two different retrieval methods, namely, BM25 and Jaccard, in Table \ref{re} and \ref{re for gen}. 
BM25 \cite{fang2004formal} is an extension of the Term Frequency-Inverse Document Frequency (TF-IDF) model, which calculates a relevance score between the query sequence and each candidate sequence in the retrieval pool. The relevance score is derived from the occurrence frequency of the nodes in the query and the retrieval pool.
Jaccard \cite{jaccard1901etude} measures the similarity between two sets by comparing the size of their intersection to the size of their union. Note that in the citation dataset Hepth and hyperlink dataset Reddit, the queries in the test set contain unseen target nodes that never appear in the retrieval pool and have no historical interactions. As a result, the BM25 and Jaccard scores between the queries and the candidates in the retrieval pool are always zeros. On the other hand, our retrieval model is trained based on the sequence representations. For a query sequence containing only the target node, we can still obtain its representation using the sequence model trained for the retrieval model, and further calculate its contextual similarity with the candidate sequences in the retrieval pool.
%use the preprocessed abstract embedding as its representation. The representation of the sequences in the retrieval pool is obtained through the retrieval model. We then calculate the dot product of the query and candidate representations for the similarity score to obtain top-\textit{K} demonstrations. 

\begin{figure*}[ht] 
\centering
    \centering
    \includegraphics[scale=0.85]{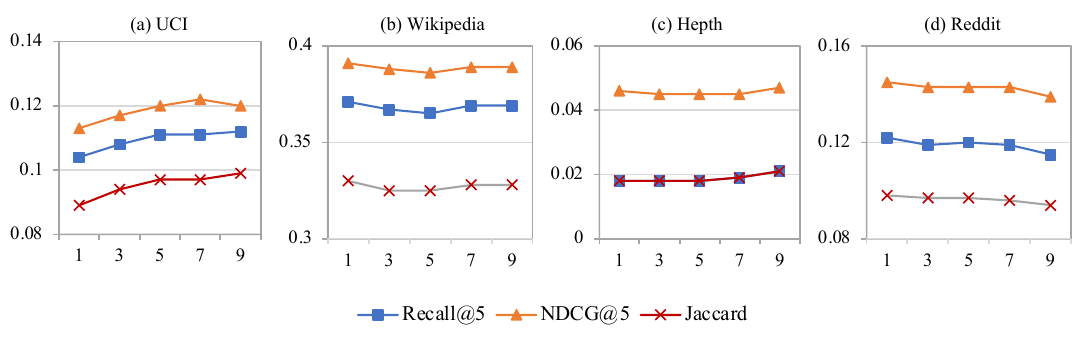}
    %\vspace{-0.1cm}
    \vspace{-0.3cm}
\caption{Effect of the number of demonstrations \textit{K}.}
\label{topK}
\end{figure*}

In Table \ref{re}, we analyze the retrieval performance of different methods. For transductive scenarios, our retrieval model shows comparable performance to other retrieval strategies. Notably, in inductive scenarios like the Hepth and Reddit datasets, BM25 and Jaccard fail to work with new query nodes lacking historical interactions. In contrast, our model can handle them effectively and achieve solid performance.

%Our retrieval model shows better performance than other retrieval strategies. Notably, in inductive scenarios like the Hepth \textcolor{blue}{and Reddit datasets}, BM25 and Jaccard fail to work with new query nodes lacking historical interactions. In contrast, our model can handle them effectively and achieve solid performance. The high $HR@3$ performance of BM25 and Jaccard on the MMConv dataset can be attributed to the nature of the dialogue dataset, where temporal order is less critical, and certain nodes associated with specific slot values are more discriminative.

Table \ref{re for gen} shows the generative performance of different retrieval methods in the dynamic link prediction task.  During testing, we apply the retrieval results obtained from different retrieval methods. We also train a model using the ground-truth retrieval results for a more comprehensive comparison. The ``GroundTruth'' row represents an upper bound on the performance when using ground-truth retrieval results on the testing data, which, as expected, provides the highest performance metrics. Generally speaking, all retrieval methods show better performance compared to the backbone SimpleDyG without using RAG, demonstrating the effectiveness of the RAG technique for dynamic graph modeling. Our method performs better compared to other retrieval strategies, indicating the effectiveness of contrastive learning in the retrieval model.

\subsubsection{Effect of the Number of Demonstrations \textit{K}}

To investigate the influence of the number of demonstrations, we conduct experiments across varying values $K \in \{1,3,5,7,9\}$. As shown in Fig.~\ref{topK}, a higher number of $K$ yields better prediction performance, that's because more demonstrations provide richer contextual information, especially in the UCI dataset. However, including too many cases may introduce more noise, which can harm the performance. Ultimately, we select $K=7$ for all datasets.

\begin{table*}[ht]
	\centering
    %\vspace{-0.1cm}
     \caption{Effect of different fusion strategies.}
    \vspace{-0.2cm}
	\small
	\addtolength{\tabcolsep}{-3pt}
	\renewcommand*{\arraystretch}{1.3}
    %\begin{tabular}{@{}c|ccc|ccc|ccc|ccc|ccc|ccc@{}}
    \begin{tabular}{@{}c|ccc|ccc|ccc|ccc@{}}
		
		\toprule
		%\multicolumn{1}{c|}{\multirow{2}{*}{Fusion strategy}}& \mc{3}{c|}{UCI} & \mc{3}{c|}{Hepth} & \mc{3}{c}{MMConv} \\ \cmidrule{2-10}
  	\multicolumn{1}{c|}{\multirow{2}{*}{Fusion strategy}}
    & \mc{3}{c|}{UCI} 
    & \mc{3}{c|}{Wikipedia}
    & \mc{3}{c|}{Hepth} 
    %& \mc{3}{c|}{MMConv}
    %& \mc{3}{c|}{Enron}
    & \mc{3}{c}{Reddit} 
    \\ \cmidrule{2-13}
		%& {Recall@5} & {NDCG@5}     & {Jaccard}   & {Recall@5}  &{NDCG@5} & {Jaccard}  & {Recall@5}   &{NDCG@5} & {Jaccard} \\	
	& {Recall@5}  & {NDCG@5} & {Jaccard}  
        & {Recall@5}  & {NDCG@5}  & {Jaccard}   
       & {Recall@5}  & {NDCG@5} & {Jaccard}  
        %& {R@5} & {NDCG}& {Jaccard}  
        %& {R@5}& {NDCG} & {Jaccard}
        & {Recall@5}  & {NDCG@5} & {Jaccard}\\			
		\midrule		

  \textit{Concatenation}  
        & 0.033\scriptsize{±0.019}   
        & 0.036\scriptsize{±0.018}
        & 0.029\scriptsize{±0.016}
        
        &0.210\scriptsize{±0.019} 
        &0.232\scriptsize{±0.021}
        &0.206\scriptsize{±0.019} 
        
        & 0.001\scriptsize{±0.002}  
        & 0.007\scriptsize{±0.002}  
        & 0.002\scriptsize{±0.002} 
        
        %& 0.103%\scriptsize{±0.031}  
        %& 0.110%\scriptsize{±0.030}  
        %& 0.103%\scriptsize{±0.031} 

        %&0.04%\scriptsize{±0.016}
        % & 0.057%\scriptsize{±0.022}
        %&0.033%\scriptsize{±0.013}

        &0.001\scriptsize{±0.001}
        &0.003\scriptsize{±0.003}
        &0.001\scriptsize{±0.001}
        
        \\ 
        
    \textit{MLP}  
        & 0.102\scriptsize{±0.018}  
        & 0.106\scriptsize{±0.017} 
        & 0.089\scriptsize{±0.016}

        & 0.356\scriptsize{±0.006} 
        &0.371\scriptsize{±0.009}
        & 0.321\scriptsize{±0.007}  
        
        & 0.006\scriptsize{±0.002}    
        & 0.015\scriptsize{±0.002} 
        & 0.006\scriptsize{±0.002}
        
        %& 0.156%\scriptsize{±0.006}  
        %& 0.167%\scriptsize{±0.006}   
        %& 0.153%\scriptsize{±0.006} 
        
        %& 0.094%\scriptsize{±0.003} 
       % & 0.111%\scriptsize{±0.003}
       % & 0.067%\scriptsize{±0.002}  
               
        & 0.108\scriptsize{±0.006}
       & 0.132\scriptsize{±0.005}
       & 0.090\scriptsize{±0.003}
        \\
        
    \textit{GraphFusion}  
          & \textbf{0.111}\scriptsize{±0.013} 
          & \textbf{0.122}\scriptsize{±0.014}    
          & \textbf{0.097}\scriptsize{±0.010}  

        & \textbf{0.369}\scriptsize{±0.006} 
        &\textbf{0.389}\scriptsize{±0.008} 
        &\textbf{0.328}\scriptsize{±0.007}           
         
         & \textbf{0.019}\scriptsize{±0.002}    
         & \textbf{0.045}\scriptsize{±0.003 }   
         &  \textbf{0.019}\scriptsize{±0.002} 
          
          %& \textbf{0.194}%\scriptsize{±0.005}    
          %& \textbf{0.208}%\scriptsize{±0.005}   
          %& \textbf{0.194}%\scriptsize{±0.005} 
                
       % & \textbf{0.1}%\scriptsize{±0.003} 
       % &\textbf{0.119}%\scriptsize{±0.004} 
       % & \textbf{0.071}%\scriptsize{±0.002} 
          
       & \textbf{0.119}\scriptsize{±0.006} 
       & \textbf{0.143}\scriptsize{±0.005} 
       & \textbf{0.096}\scriptsize{±0.003}         
          
          \\ 
               
		\bottomrule

	\end{tabular}
    %}
     \label{fusion}
 %\vspace{-0.2cm}    
\end{table*}

\subsubsection{Effect of Different Fusion Strategies.}

To further investigate the effectiveness of the fusion strategy for the top-\textit{K} demonstrations, we conduct experiments with different fusion strategies under \textit{K} $=7$.  ``Concatenation'' denotes we directly concatenate the sequences of retrieved demonstrations and prepend them with the query sample sequence and then feed it into the pre-trained SimpleDyG model. ``MLP'' means we do not consider the graph structure of the demonstrations and replace the graph fusion as an MLP layer (we set the number of the MLP layer as 2). By using the MLP layer, We map the concatenated demonstrations into shorter \textit{m}-dimensional embeddings (we empirically set \textit{m} to be 15) and then concatenate it with the query sample. Like graph fusion, we only fine-tune the parameters of the MLP and output layer. The results in Table \ref{fusion} show that directly concatenating the retrieved demonstrations with the query sample leads to lower performance compared with other strategies. This is because simple concatenation introduces a lengthy context, which can overwhelm the model with irrelevant information, and it neglects the structural relationships inherent in the demonstrations. The ``MLP'' strategy improves upon this by mapping the concatenated demonstrations into a shorter feature space, effectively reducing noise and emphasizing more relevant features. This approach yields better results than simple concatenation but still falls short compared to the ``GraphFusion'' strategy. The superior performance of the ``GraphFusion'' strategy highlights the importance of considering both the content and the structure of the demonstrations in the fusion process.

%\vspace{-0.2cm}
\subsubsection{Time Complexity Analysis}
\begin{figure}[h] 
\centering
    \centering
    \includegraphics[scale=0.8]{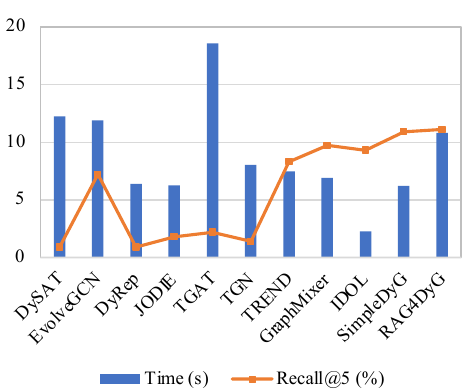}
    \vspace{-0.2cm}
\caption{Time efficiency and Recall@5 of different methods.}
    
\label{time}
\vspace{-0.5cm}
\end{figure}

The time complexity of our model RAG4DyG aligns with that of the vanilla Transformer, scaling as $O(n^2)$, where $n$ denotes the sequence length. We measured the training time per epoch across various approaches using the UCI dataset to assess its efficiency. The results shown in Table~\ref{time} indicate that our model achieves faster or comparable training cost compared to the baseline methods. In contrast, approaches that integrate temporal components (e.g., RNNs or self-attention mechanisms) with structural elements (e.g., GNNs or GATs) face significant computational challenges due to the complexity of combining these modules.

\section{Conclusion}
%In this work, we introduced a novel retrieval augmented framework RAG4DyG for dynamic graph modeling, addressing the limitations of existing approaches that often rely on narrow historical contexts. We leverage the RAG technique to broaden the context by incorporating relevant auxiliary information by harvesting high-quality demonstrations for dynamic graph samples and effectively utilizing this demonstration to enhance dynamic graph modeling. Our proposed solution involves a time-aware contrastive learning model to identify and retrieve temporally pertinent cases for each query sequence, coupled with a graph fusion strategy to integrate the inherent historical context with extended temporal contexts. Extensive experiments on real-world datasets across different domains demonstrated the superior performance of RAG4DyG in dynamic graph modeling.
In this work, we proposed RAG4DyG, a novel retrieval-augmented framework for dynamic graph modeling that overcomes the limitations of existing approaches by integrating broader temporal and contextual information. By leveraging the retrieval-augmented generation paradigm, RAG4DyG retrieves high-quality demonstrations from a retrieval pool and incorporates them effectively into the modeling process. The framework includes a time-aware contrastive learning module to prioritize temporally relevant samples and a graph fusion strategy to seamlessly integrate these retrieved demonstrations with the query sequence, enriching the historical context with extended temporal insights. Extensive experiments on diverse real-world datasets demonstrate the effectiveness of RAG4DyG in achieving state-of-the-art performance for dynamic graph modeling in both transductive and inductive scenarios.

%%
%% The acknowledgments section is defined using the "acks" environment
%% (and NOT an unnumbered section). This ensures the proper
%% identification of the section in the article metadata, and the
%% consistent spelling of the heading.
\begin{acks}
%To Robert, for the bagels and explaining CMYK and color spaces.
This research / project is supported by the Ministry of Education, Singapore, under its Academic Research Fund Tier 2 (Proposal ID: T2EP20122-0041 and T2EP20123-0052). Any opinions, findings and conclusions or recommendations expressed in this material are those of the author(s) and
do not reflect the views of the Ministry of Education, Singapore. The authors would also like to thank HOANG Thi Linh for her assistance with this work. 
\end{acks}

%%
%% The next two lines define the bibliography style to be used, and
%% the bibliography file.

%\appendix
%\input{app}
\balance

\bibliographystyle{ACM-Reference-Format}
\bibliography{ref}

%%
%% If your work has an appendix, this is the place to put it.

%\clearpage

\end{document}